# Bayesian Efficient Multiple Kernel Learning

**Mehmet Gönen**                                                                          MEHMET.GONEN@AALTO.FI

Helsinki Institute for Information Technology HIIT
Department of Information and Computer Science, Aalto University School of Science

## Abstract

Multiple kernel learning algorithms are proposed to combine kernels in order to obtain a better similarity measure or to integrate feature representations coming from different data sources. Most of the previous research on such methods is focused on the computational efficiency issue. However, it is still not feasible to combine many kernels using existing Bayesian approaches due to their high time complexity. We propose a fully conjugate Bayesian formulation and derive a deterministic variational approximation, which allows us to combine hundreds or thousands of kernels very efficiently. We briefly explain how the proposed method can be extended for multiclass learning and semi-supervised learning. Experiments with large numbers of kernels on benchmark data sets show that our inference method is quite fast, requiring less than a minute. On one bioinformatics and three image recognition data sets, our method outperforms previously reported results with better generalization performance.

## 1. Introduction

The main idea of kernel-based algorithms is to learn a linear decision function in the feature space where data points are implicitly mapped to using a kernel function (Vapnik, 1998). Given a sample of $N$ independent and identically distributed training instances $\{\boldsymbol{x}_i \in \mathcal{X}\}_{i=1}^N$, the decision function that is used to predict the target output of an unseen test instance $\boldsymbol{x}_\star$ can be written as

$$f(\boldsymbol{x}_\star) = \boldsymbol{a}^\top \boldsymbol{k}_\star + b \qquad (1)$$

where the vector of weights assigned to each training data point and the bias are denoted by $\boldsymbol{a}$ and $b$, respectively, and $\boldsymbol{k}_\star = \begin{bmatrix} k(\boldsymbol{x}_1, \boldsymbol{x}_\star) & \ldots & k(\boldsymbol{x}_N, \boldsymbol{x}_\star) \end{bmatrix}^\top$ where $k \colon \mathcal{X} \times \mathcal{X} \to \mathbb{R}$ is the kernel function that calculates a similarity measure between two data points. Using the theory of structural risk minimization, the model parameters can be found by solving a quadratic programming problem, known as *support vector machine* (SVM) (Vapnik, 1998). The model parameters can also be interpreted as random variables to obtain a Bayesian interpretation of the model, known as *relevance vector machine* (RVM) (Tipping, 2001).

Kernel selection (i.e., choosing a functional form and its parameters) is the most important issue that affects the empirical performance of kernel-based algorithms and is usually done using a cross-validation procedure. *Multiple kernel learning* (MKL) methods have been proposed to make use of multiple kernels simultaneously instead of selecting a single kernel (see a recent survey by Gönen & Alpaydın (2011)). Such methods also provide a principled way of integrating feature representations coming from different data sources or modalities. Most of the previous research is focused on developing efficient MKL algorithms. Nevertheless, existing Bayesian MKL methods are problematic in terms of computation time when combining hundreds or thousands of kernels. In this paper, we formulate a very efficient Bayesian MKL method that solves this issue by formulating the combination in a novel way.

In Section 2, we give an overview of the related work by considering existing discriminative and Bayesian MKL algorithms. Section 3 gives the details of the proposed fully conjugate Bayesian formulation, called *Bayesian efficient multiple kernel learning* (BEMKL). In Section 4, we explain detailed derivations of our deterministic variational approximation for binary classification. Extensions towards multiclass learning and semi-supervised learning are summarized in Section 5. Section 6 evaluates BEMKL with large numbers of kernels on standard benchmark data sets in terms of time complexity, and reports the classification results on one bioinformatics and three image recognition tasks, which are frequently used to compare MKL methods.

## 2. Related Work

MKL algorithms basically replace the kernel in (1) with a combined kernel calculated as a function of the input kernels. The most common combination is to use a weighted sum of $P$ kernels $\{k_m \colon \mathcal{X} \times \mathcal{X} \to \mathbb{R}\}_{m=1}^{P}$:

$$f(\boldsymbol{x}_\star) = \boldsymbol{a}^\top \underbrace{\left(\sum_{m=1}^{P} e_m \boldsymbol{k}_{m,\star}\right)}_{\boldsymbol{k}_{e,\star}} + b$$

where the vector of kernel weights is denoted by $\boldsymbol{e}$ and $\boldsymbol{k}_{m,\star} = \begin{bmatrix} k_m(\boldsymbol{x}_1, \boldsymbol{x}_\star) & \ldots & k_m(\boldsymbol{x}_N, \boldsymbol{x}_\star) \end{bmatrix}^\top$. Existing MKL algorithms with a weighted sum differ in the way that they formulate restrictions on the kernel weights: arbitrary weights (i.e., linear sum), nonnegative weights (i.e., conic sum), or weights on a simplex (i.e., convex sum).

Bach et al. (2004) formulate the problem as a *second-order cone programming* (SCOP) problem, which is formulated as a semidefinite programming problem previously by Lanckriet et al. (2004). However, SCOP can only be solved for medium-scale problems efficiently. Sonnenburg et al. (2006) reinterpret the problem as a *semi-infinite linear programming* (SILP) problem, which can be applied to large-scale data sets. Rakotomamonjy et al. (2008) develop a simple MKL algorithm using a *sub-gradient descent* (SD) approach, which is faster than SILP method. Later, Xu et al. (2009) extend the *level method*, which is originally designed for optimizing non-smooth objective functions, to obtain a very efficient MKL algorithm that carries flavors from both SILP and SD approaches but outperforms them in terms of computation time.

The aforementioned methods tend to produce sparse kernel combinations, which corresponds to using the $\ell_1$-norm on the kernel weights. Sparsity at the kernel level may harm the generalization performance of the learner and using non-sparse kernel combinations (e.g., the $\ell_2$-norm) may be a better choice (Cortes et al., 2009). Varma & Babu (2009) propose a generalized MKL algorithm that can use any differentiable and continuous regularization term on the kernel weights. This also allows us to integrate prior knowledge about the kernels to the model. Xu et al. (2010) and Kloft et al. (2011) independently and in parallel develop an MKL algorithm with the $\ell_p$-norm ($p \geq 1$) on the kernel weights. This method has a closed-form update rule for the kernel weights and requires only an SVM solver for optimization. *Sequential minimal optimization* (SMO) algorithm is the most commonly used method for solving SVM problems and efficiently scales to large problems. Vishwanathan et al. (2010) propose a very efficient method, called SMO-MKL, to train $\ell_p$-norm ($p > 1$) MKL models with squared norm as the regularization term using SMO algorithm for solving MKL problem directly instead of solving intermediate SVMs at each iteration.

Most of the discriminative MKL algorithms are developed for binary classification. One-versus-all or one-versus-other strategies can be employed to get multiclass learners. However, there are also some direct formulations for multiclass learning. Zien & Ong (2007) give a multiclass MKL algorithm by formulating the problem as an SILP and show that their method is equivalent to multiclass generalizations of Bach et al. (2004) and Sonnenburg et al. (2006). Gehler & Nowozin (2009) propose a boosting-type MKL algorithm that combines outputs calculated from each kernel separately and obtain better results than MKL algorithms with SILP and SD approaches on image recognition problems.

Girolami & Rogers (2005) present Bayesian MKL algorithms for regression and binary classification using hierarchical models. Damoulas & Girolami (2008) give a multiclass MKL formulation using a very similar hierarchical model. The combined kernel in these two studies is defined as a convex sum of the input kernels using a Dirichlet prior on the kernel weights. As a consequence of the nonconjugacy between Dirichlet and normal distributions, they choose to use an importance sampling scheme to update the kernel weights when deriving variational approximations. Recently, Zhang et al. (2011) propose a fully Bayesian inference methodology for extending generalized linear models to kernelized models using a *Markov chain Monte Carlo* (MCMC) approach. The main issue with these approaches is that they depend on some sampling strategy and may not be trained in a reasonable time when the number of kernels is large.

Girolami & Zhong (2007) formulate a *Gaussian process* (GP) variant that uses multiple covariances (i.e., kernels) for multiclass classification using a variational approximation or expectation propagation scheme, which requires an MCMC sub-sampler for the covariance weights. Titsias & Lázaro-Gredilla (2011) propose a multitask GP model that combines a common set of GP functions (i.e., information sharing between the tasks) defined over multiple covariances with task-dependent weights whose sparsity is tuned using the spike and slab prior. A variational approximation approach is derived for an efficient inference scheme.

Our main motivation for this work is to formulate an efficient Bayesian inference approach without resorting to expensive sampling procedures.

## 3. Bayesian Efficient Multiple Kernel Learning

In order to obtain an efficient Bayesian MKL algorithm, we formulate a fully conjugate probabilistic model and develop a deterministic variational approximation mechanism for inference. We give the details for binary classification, but the same model can easily be extended to regression.

Figure 1 illustrates the proposed probabilistic model for binary classification with a graphical model. The main idea is to calculate intermediate outputs from each kernel using the same set of weight parameters and to combine these outputs using the kernel weights and the bias to estimate the classification score. Different from earlier probabilistic MKL approaches such as Girolami & Rogers (2005) and Damoulas & Girolami (2008), our method has two key properties that enable us to perform efficient inference: (a) Intermediate outputs calculated using the input kernels are introduced. (b) Kernel weights are assumed to be normally distributed without enforcing any constraints on them.

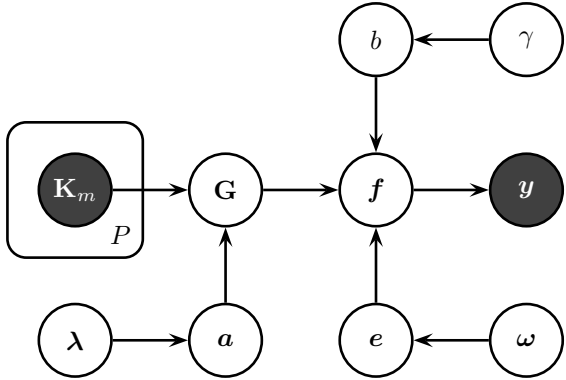

*Figure 1.* Bayesian efficient MKL for binary classification.

The notation we use throughout the rest of the manuscript is as follows: $N$ and $P$ represent the numbers of training instances and input kernels, respectively. The $N \times N$ kernel matrices are denoted by $\mathbf{K}_m$, where the columns of $\mathbf{K}_m$ by $\boldsymbol{k}_{m,i}$ and the rows of $\mathbf{K}_m$ by $\boldsymbol{k}_m^i$. The $N \times 1$ vectors of weight parameters $a_i$ and their priors $\lambda_i$ are denoted by $\boldsymbol{a}$ and $\boldsymbol{\lambda}$, respectively. The $P \times N$ matrix of intermediate outputs $g_i^m$ is represented as $\mathbf{G}$, where the columns of $\mathbf{G}$ as $\boldsymbol{g}_i$ and the rows of $\mathbf{G}$ as $\boldsymbol{g}^m$. The bias parameter and its prior are denoted by $b$ and $\gamma$, respectively. The $P \times 1$ vectors of kernel weights $e_m$ and their priors $\omega_m$ are denoted by $\boldsymbol{e}$ and $\boldsymbol{\omega}$, respectively. The $N \times 1$ vector of auxiliary variables $f_i$ is represented as $\boldsymbol{f}$. The $N \times 1$ vector of associated class labels is represented as $\boldsymbol{y}$, where each element $y_i \in \{-1, +1\}$.

As short-hand notations, all priors in the model are denoted by $\boldsymbol{\Xi} = \{\gamma, \boldsymbol{\lambda}, \boldsymbol{\omega}\}$, where the remaining variables by $\boldsymbol{\Theta} = \{\boldsymbol{a}, b, \boldsymbol{e}, \boldsymbol{f}, \mathbf{G}\}$ and the hyper-parameters by $\boldsymbol{\zeta} = \{\alpha_\gamma, \beta_\gamma, \alpha_\lambda, \beta_\lambda, \alpha_\omega, \beta_\omega\}$. Dependence on $\boldsymbol{\zeta}$ is omitted for clarity throughout the manuscript.

The distributional assumptions of our proposed model are defined as

$$\begin{aligned}
\lambda_i &\sim \mathcal{G}(\lambda_i; \alpha_\lambda, \beta_\lambda) & \forall i \\
a_i | \lambda_i &\sim \mathcal{N}(a_i; 0, \lambda_i^{-1}) & \forall i \\
g_i^m | \boldsymbol{a}, \boldsymbol{k}_{m,i} &\sim \mathcal{N}(g_i^m; \boldsymbol{a}^\top \boldsymbol{k}_{m,i}, 1) & \forall (m, i) \\
\gamma &\sim \mathcal{G}(\gamma; \alpha_\gamma, \beta_\gamma) \\
b | \gamma &\sim \mathcal{N}(b; 0, \gamma^{-1}) \\
\omega_m &\sim \mathcal{G}(\omega_m; \alpha_\omega, \beta_\omega) & \forall m \\
e_m | \omega_m &\sim \mathcal{N}(e_m; 0, \omega_m^{-1}) & \forall m \\
f_i | b, \boldsymbol{e}, \boldsymbol{g}_i &\sim \mathcal{N}(f_i; \boldsymbol{e}^\top \boldsymbol{g}_i + b, 1) & \forall i \\
y_i | f_i &\sim \delta(f_i y_i > \nu) & \forall i
\end{aligned}$$

where the auxiliary variables between the intermediate outputs and the class labels are introduced to make the inference procedures efficient (Albert & Chib, 1993), and the margin parameter $\nu$ is introduced to resolve the scaling ambiguity issue and to place a low-density region between two classes, similar to the margin idea in SVMs, which is generally used for semi-supervised learning (Lawrence & Jordan, 2005). $\mathcal{N}(\cdot; \boldsymbol{\mu}, \boldsymbol{\Sigma})$ represents the normal distribution with the mean vector $\boldsymbol{\mu}$ and the covariance matrix $\boldsymbol{\Sigma}$. $\mathcal{G}(\cdot; \alpha, \beta)$ denotes the gamma distribution with the shape parameter $\alpha$ and the scale parameter $\beta$. $\delta(\cdot)$ represents the Kronecker delta function that returns 1 if its argument is true and 0 otherwise.

When we consider the random variables in our model as deterministic values, the auxiliary variable that corresponds to the decision function value in discriminative methods can be decomposed as

$$f_\star = \boldsymbol{e}^\top \boldsymbol{g}_\star + b = \sum_{m=1}^P e_m g_\star^m + b = \sum_{m=1}^P e_m \boldsymbol{a}^\top \boldsymbol{k}_{m,\star} + b$$
$$= \boldsymbol{a}^\top \left( \sum_{m=1}^P e_m \boldsymbol{k}_{m,\star} \right) + b = \boldsymbol{a}^\top \boldsymbol{k}_{e,\star} + b$$

where we see that the combined kernel in our model can be expressed as a linear sum of the input kernels.

Note that sample-level sparsity can be tuned by assigning suitable values to the hyper-parameters $(\alpha_\lambda, \beta_\lambda)$ as in RVMs (Tipping, 2001). Kernel-level sparsity can also be tuned by changing the hyper-parameters $(\alpha_\omega, \beta_\omega)$. Sparsity-inducing gamma priors can simulate the $\ell_1$-norm on the kernel weights, whereas uninformative priors simulate the $\ell_2$-norm.

## 4. Efficient Inference Using Variational Approximation

Exact inference for our probabilistic model is intractable and using a Gibbs sampling approach is computationally expensive (Gelfand & Smith, 1990). We instead formulate a deterministic variational approximation, which is more efficient in terms of computation time. The variational methods use a lower bound on the marginal likelihood using an ensemble of factored posteriors to find the joint parameter distribution (Beal, 2003). We can write the factorable ensemble approximation of the required posterior as

$$p(\Theta, \Xi | \{\mathbf{K}_m\}_{m=1}^P, \boldsymbol{y}) \approx q(\Theta, \Xi) = $$
$$q(\boldsymbol{\lambda})q(\boldsymbol{a})q(\mathbf{G})q(\gamma)q(\boldsymbol{\omega})q(b,\boldsymbol{e})q(\boldsymbol{f})$$

and define each factor in the ensemble just like its full conditional distribution:

$$q(\boldsymbol{\lambda}) = \prod_{i=1}^{N} \mathcal{G}(\lambda_i; \alpha(\lambda_i), \beta(\lambda_i))$$
$$q(\boldsymbol{a}) = \mathcal{N}(\boldsymbol{a}; \mu(\boldsymbol{a}), \Sigma(\boldsymbol{a}))$$
$$q(\mathbf{G}) = \prod_{i=1}^{N} \mathcal{N}(\boldsymbol{g}_i; \mu(\boldsymbol{g}_i), \Sigma(\boldsymbol{g}_i))$$
$$q(\gamma) = \mathcal{G}(\gamma; \alpha(\gamma), \beta(\gamma))$$
$$q(\boldsymbol{\omega}) = \prod_{m=1}^{P} \mathcal{G}(\omega_m; \alpha(\omega_m), \beta(\omega_m))$$
$$q(b,\boldsymbol{e}) = \mathcal{N}\left(\begin{bmatrix}b\\\boldsymbol{e}\end{bmatrix}; \mu(b,\boldsymbol{e}), \Sigma(b,\boldsymbol{e})\right)$$
$$q(\boldsymbol{f}) = \prod_{i=1}^{N} \mathcal{TN}(f_i; \mu(f_i), \Sigma(f_i), \rho(f_i))$$

where $\alpha(\cdot)$, $\beta(\cdot)$, $\mu(\cdot)$, and $\Sigma(\cdot)$ denote the shape parameter, the scale parameter, the mean vector, and the covariance matrix for their arguments, respectively. $\mathcal{TN}(\cdot; \boldsymbol{\mu}, \boldsymbol{\Sigma}, \rho(\cdot))$ denotes the truncated normal distribution with the mean vector $\boldsymbol{\mu}$, the covariance matrix $\boldsymbol{\Sigma}$, and the truncation rule $\rho(\cdot)$ such that $\mathcal{TN}(\cdot; \boldsymbol{\mu}, \boldsymbol{\Sigma}, \rho(\cdot)) \propto \mathcal{N}(\cdot; \boldsymbol{\mu}, \boldsymbol{\Sigma})$ if $\rho(\cdot)$ is true and $\mathcal{TN}(\cdot; \boldsymbol{\mu}, \boldsymbol{\Sigma}, \rho(\cdot)) = 0$ otherwise.

We can bound the marginal likelihood using Jensen's inequality:

$$\log p(\boldsymbol{y}|\{\mathbf{K}_m\}_{m=1}^P) \geq$$
$$\mathrm{E}_{q(\Theta,\Xi)}[\log p(\boldsymbol{y},\Theta,\Xi|\{\mathbf{K}_m\}_{m=1}^P)]$$
$$- \mathrm{E}_{q(\Theta,\Xi)}[\log q(\Theta,\Xi)] \quad (2)$$

and optimize this bound by maximizing with respect to each factor separately until convergence. The approximate posterior distribution of a specific factor $\boldsymbol{\tau}$ can be found as

$$q(\boldsymbol{\tau}) \propto \exp(\mathrm{E}_{q(\{\Theta,\Xi\}\setminus\boldsymbol{\tau})}[\log p(\boldsymbol{y},\Theta,\Xi|\{\mathbf{K}_m\}_{m=1}^P)]).$$

For our model, thanks to the conjugacy, the resulting approximate posterior distribution of each factor follows the same distribution as the corresponding factor.

### 4.1. Inference Details

The approximate posterior distributions of the precision priors can be found as

$$q(\boldsymbol{\lambda}) = \prod_{i=1}^{N} \mathcal{G}\left(\lambda_i; \alpha_\lambda + \frac{1}{2}, \left(\frac{1}{\beta_\lambda} + \frac{\widetilde{a_i^2}}{2}\right)^{-1}\right)$$

$$q(\gamma) = \mathcal{G}\left(\gamma; \alpha_\gamma + \frac{1}{2}, \left(\frac{1}{\beta_\gamma} + \frac{\widetilde{b^2}}{2}\right)^{-1}\right)$$

$$q(\boldsymbol{\omega}) = \prod_{m=1}^{P} \mathcal{G}\left(\omega_m; \alpha_\omega + \frac{1}{2}, \left(\frac{1}{\beta_\omega} + \frac{\widetilde{e_m^2}}{2}\right)^{-1}\right)$$

where the tilde notation denotes the posterior expectations as usual, i.e., $\widetilde{h(\boldsymbol{\tau})} = \mathrm{E}_{q(\boldsymbol{\tau})}[h(\boldsymbol{\tau})]$.

The approximate posterior distribution of the weight parameters is a multivariate normal distribution:

$$q(\boldsymbol{a}) = \mathcal{N}\left(\boldsymbol{a}; \Sigma(\boldsymbol{a})\left(\sum_{m=1}^{P} \mathbf{K}_m \widetilde{(\boldsymbol{g}^m)^\top}\right),\right.$$
$$\left.\left(\mathrm{diag}(\widetilde{\boldsymbol{\lambda}}) + \sum_{m=1}^{P} \mathbf{K}_m \mathbf{K}_m^\top\right)^{-1}\right). \quad (3)$$

The approximate posterior distribution of the intermediate outputs can be found as a product of multivariate normal distributions:

$$q(\mathbf{G}) = \prod_{i=1}^{N} \mathcal{N}\left(\boldsymbol{g}_i; \Sigma(\boldsymbol{g}_i)\left(\begin{bmatrix}\boldsymbol{k}_1^i\\\vdots\\\boldsymbol{k}_P^i\end{bmatrix}\widetilde{\boldsymbol{a}} + \widetilde{f_i}\widetilde{\boldsymbol{e}} - \widetilde{b}\widetilde{\boldsymbol{e}}\right),\right.$$
$$\left.\left(\mathbf{I} + \widetilde{\boldsymbol{e}\boldsymbol{e}^\top}\right)^{-1}\right).$$

The approximate posterior distribution of the bias and the kernel weights can be formulated as a multivariate normal distribution:

$$q(b,\boldsymbol{e}) = \mathcal{N}\left(\begin{bmatrix}b\\\boldsymbol{e}\end{bmatrix}; \Sigma(b,\boldsymbol{e})\begin{bmatrix}\mathbf{1}^\top \widetilde{\boldsymbol{f}}\\\widetilde{\mathbf{G}}\widetilde{\boldsymbol{f}}\end{bmatrix},\right.$$
$$\left.\begin{bmatrix}\widetilde{\gamma}+N & \mathbf{1}^\top\widetilde{\mathbf{G}^\top}\\\widetilde{\mathbf{G}}\mathbf{1} & \mathrm{diag}(\widetilde{\boldsymbol{\omega}}) + \widetilde{\mathbf{G}\mathbf{G}^\top}\end{bmatrix}^{-1}\right) \quad (4)$$

where we allow kernel weights to take negative values. If the kernel weights are restricted to be nonnegative, the approximation becomes a truncated multivariate distribution restricted to the nonnegative orthant except for the first dimension, which is used for the bias.

The approximate posterior distribution of the auxiliary variables is a product of truncated normal distributions given as

$$q(\boldsymbol{f}) = \prod_{i=1}^{N} \mathcal{TN}(f_i; \widetilde{\boldsymbol{e}}^\top \widetilde{\boldsymbol{g}}_i + \widetilde{b}, 1, f_i y_i > \nu)$$

where we need to find their posterior expectations to update the approximate posterior distributions of the intermediate outputs, the bias, and the kernel weights. Fortunately, the truncated normal distribution has a closed-form formula for its expectation.

$\sum_{m=1}^{P} \mathbf{K}_m \mathbf{K}_m^\top$ in (3) should be cached before starting inference to reduce the computational complexity. (3) requires inverting an $N \times N$ matrix for the covariance calculation, whereas (4) requires inverting a $(P+1) \times (P+1)$ matrix. One of these two updates dominates the running time depending on whether $N > P$.

### 4.2. Convergence

The inference mechanism sequentially updates the approximate posterior distributions of the model parameters and the latent variables until convergence, which can be checked by monitoring the lower bound in (2). The first term of the lower bound corresponds to the sum of exponential form expectations of the distributions in the joint likelihood. The second term is the sum of negative entropies of the approximate posteriors in the ensemble. The only nonstandard distribution in these terms is the truncated normal distribution used for the auxiliary variables; nevertheless, the truncated normal distribution has a closed-form formula also for its entropy. Exact form of the variational lower bound can be found in the supplementary material available at http://users.ics.aalto.fi/gonen/bemkl/.

### 4.3. Prediction

We can replace $p(\boldsymbol{a}|\{\mathbf{K}_m\}_{m=1}^{P}, \boldsymbol{y})$ with its approximate posterior distribution $q(\boldsymbol{a})$ and obtain the predictive distribution of the intermediate outputs $\boldsymbol{g}_\star$ for a new data point as

$$p(\boldsymbol{g}_\star|\{\boldsymbol{k}_{m,\star}, \mathbf{K}_m\}_{m=1}^{P}, \boldsymbol{y}) = \prod_{m=1}^{P} \mathcal{N}(g_\star^m; \mu(\boldsymbol{a})^\top \boldsymbol{k}_{m,\star}, 1 + \boldsymbol{k}_{m,\star}^\top \Sigma(\boldsymbol{a}) \boldsymbol{k}_{m,\star}).$$

The predictive distribution of the auxiliary variable $f_\star$ can also be found by replacing $p(b, \boldsymbol{e}|\{\mathbf{K}_m\}_{m=1}^{P}, \boldsymbol{y})$ with its approximate posterior distribution $q(b, \boldsymbol{e})$:

$$p(f_\star|\boldsymbol{g}_\star, \{\mathbf{K}_m\}_{m=1}^{P}, \boldsymbol{y}) = \\ \mathcal{N}\left(f_\star; \mu(b, \boldsymbol{e})^\top \begin{bmatrix} 1 \\ \boldsymbol{g}_\star \end{bmatrix}, 1 + \begin{bmatrix} 1 & \boldsymbol{g}_\star \end{bmatrix} \Sigma(b, \boldsymbol{e}) \begin{bmatrix} 1 \\ \boldsymbol{g}_\star \end{bmatrix}\right)$$

and the predictive distribution of the class label $y_\star$ can be formulated using the auxiliary variable distribution:

$$p(y_\star = +1|\{\boldsymbol{k}_{m,\star}, \mathbf{K}_m\}_{m=1}^{P}, \boldsymbol{y}) = \mathcal{Z}_\star^{-1} \Phi\left(\frac{\mu(f_\star) - \nu}{\Sigma(f_\star)}\right)$$

where $\mathcal{Z}_\star$ is the normalization coefficient calculated for the test data point and $\Phi(\cdot)$ is the standardized normal cumulative distribution function.

## 5. Extensions

The proposed MKL method can be extended in several directions. We shortly explain two variants for multiclass learning and semi-supervised learning.

### 5.1. Multiclass Learning

In multiclass learning, we consider classification problems with more than two classes. The easiest way is to train a distinct classifier for each class that separates this particular class from the remaining (i.e., one-versus-all classification). In this setup, for each classifier, we learn a different combined kernel function, which corresponds to using a different similarity measure. Instead, we can have different classifiers but use the same set of kernel weights for each of them (Rakotomamonjy et al., 2008). The inference details of our model with such an approach can be found in the supplementary material. Another possibility is to use multinomial probit in our model for multiclass classification as done by Damoulas & Girolami (2008).

### 5.2. Semi-Supervised Learning

In kernel-based discriminative learning framework, semi-supervised learning can be formulated as an integer programming problem, known as transductive SVMs (Vapnik, 1998). Even though there are some approximation methods, the computational complexity of such models are significantly higher than that of fully supervised models. Considering this complexity issue, it may not be feasible to use discriminative MKL algorithms for semi-supervised learning. However, our proposed model can be modified to make use of unlabeled data with a slight increase in computational complexity using the low-density assumption (Lawrence & Jordan, 2005).

## 6. Experiments

We first test our new algorithm BEMKL on benchmark data sets to show its computational efficiency. We then illustrate its generalization performance comparing it with previously reported MKL results on one bioinformatics and three image recognition data sets. We implement the proposed variational approximation for BEMKL in Matlab and our implementation is available at http://users.ics.aalto.fi/gonen/bemkl/.

### 6.1. Experiments on Benchmark Data Sets

Our first set of experiments is designed to evaluate the running times of BEMKL. The experiments are performed on eight data sets from the UCI repository: `breast`, `bupaliver`, `heart`, `ionosphere`, `pima`, `sonar`, `wdbc`, and `wpbc`. For each data set, we take 20 replications, where we randomly select 70 per cent of the data set as the training set and use the remaining as the test set. The training set is normalized to have zero mean and unit standard deviation, and the test set is then normalized using the mean and the standard deviation of the training set.

We construct Gaussian kernels with 10 different widths ($\{2^{-3}, 2^{-2}, \ldots, 2^{6}\}$) and polynomial kernels with three different degrees ($\{1, 2, 3\}$) on all features and on each single feature, following the experimental settings of Rakotomamonjy et al. (2008), Xu et al. (2009; 2010), and Vishwanathan et al. (2010). All kernel matrices are normalized to have unit diagonal entries (i.e., spherical normalization) and precomputed before running the inference algorithm. The experiments are performed on a PC with 3.0GHz CPU and 4GB memory. We run BEMKL for two different sparsity levels on the kernel weights: sparse and non-sparse. The hyperparameter values for these scenarios are selected as $(\alpha_\lambda, \beta_\lambda, \alpha_\gamma, \beta_\gamma, \alpha_\omega, \beta_\omega) = (1, 1, 1, 1, 10^{-10}, 10^{+10})$ and $(\alpha_\lambda, \beta_\lambda, \alpha_\gamma, \beta_\gamma, \alpha_\omega, \beta_\omega) = (1, 1, 1, 1, 1, 1)$, respectively. We take 200 iterations for variational inference scheme.

Table 1 lists the results obtained by BEMKL with two scenarios on benchmark data sets in terms of three different measures: the training time in seconds, the test accuracy, and the number of selected kernels. We see that our deterministic variational approximation for BEMKL takes less than a minute with large numbers of kernels, ranging from 91 to 793. The classification accuracy difference between sparse and non-sparse kernel combination is quite obvious for some data sets such as `sonar` in accordance with the previous studies. Using sparsity inducing priors on kernel weights clearly simulates the $\ell_1$-norm by eliminating most of the input kernels, whereas using uninformative priors picks much more kernels by simulating the $\ell_2$-norm.

Table 1. Experiments on benchmark data sets. The figures are averages and standard deviations over 20 replications. Here, $N$ and $P$ denote the numbers of training instances and input kernels, respectively.

|  | sparse | non-sparse |
|---|---|---|
| `breast` | $N = 478$ | $P = 130$ |
| Training Time (sec) | 18.86±0.24 | 18.84±0.43 |
| Test Accuracy (%) | 96.80±1.02 | 96.98±0.86 |
| Selected Kernel (#) | 34.35±2.46 | 98.95±1.32 |
| `bupaliver` | $N = 241$ | $P = 91$ |
| Training Time (sec) | 3.83±0.03 | 3.81±0.04 |
| Test Accuracy (%) | 68.41±4.28 | 70.72±3.83 |
| Selected Kernel (#) | 18.40±2.62 | 70.70±1.13 |
| `heart` | $N = 189$ | $P = 377$ |
| Training Time (sec) | 13.69±0.08 | 13.58±0.09 |
| Test Accuracy (%) | 80.80±4.01 | 81.36±4.70 |
| Selected Kernel (#) | 41.85±4.26 | 181.40±8.93 |
| `ionosphere` | $N = 245$ | $P = 442$ |
| Training Time (sec) | 22.72±0.05 | 22.76±0.12 |
| Test Accuracy (%) | 92.03±1.72 | 92.03±1.77 |
| Selected Kernel (#) | 41.90±4.42 | 219.05±7.77 |
| `pima` | $N = 537$ | $P = 117$ |
| Training Time (sec) | 21.15±0.23 | 20.94±0.22 |
| Test Accuracy (%) | 75.02±2.28 | 74.96±2.08 |
| Selected Kernel (#) | 23.20±2.02 | 79.55±2.93 |
| `sonar` | $N = 144$ | $P = 793$ |
| Training Time (sec) | 51.34±0.19 | 51.51±0.10 |
| Test Accuracy (%) | 76.88±3.95 | 82.81±4.12 |
| Selected Kernel (#) | 15.30±3.40 | 372.80±7.78 |
| `wdbc` | $N = 398$ | $P = 403$ |
| Training Time (sec) | 41.31±0.21 | 41.44±0.16 |
| Test Accuracy (%) | 95.70±2.32 | 95.76±1.65 |
| Selected Kernel (#) | 35.65±2.28 | 215.50±4.81 |
| `wpbc` | $N = 135$ | $P = 429$ |
| Training Time (sec) | 12.72±0.11 | 12.67±0.23 |
| Test Accuracy (%) | 75.25±1.01 | 73.64±2.66 |
| Selected Kernel (#) | 28.25±2.75 | 220.10±9.83 |

### 6.2. Comparison on MKL Data Sets

We use four data sets that are previously used to compare MKL algorithms and have kernel matrices available for direct evaluation. Note that the results used for comparison may not be the best results reported on these data sets but we use the exact same kernel matrices that produce these results for our algorithm to have comparable performance measures. These data sets have small numbers of kernels available and we do not force any sparsity on the kernel weights using an uninformative Gamma prior in accordance with the previous studies. All data sets consider multiclass classification problems and we report both one-versus-all and multiclass results for BEMKL.

Table 2. Performance comparison on `Protein` data set.

| Method | Test Accuracy |
| --- | --- |
| Damoulas & Girolami (2008) | 68.1±1.2 |
| BEMKL (one-versus-all) | 71.5±0.1 |
| BEMKL (multiclass) | 71.2±0.2 |

Table 3. Performance comparison on `Flowers17` data set.

| Method | Test Accuracy |
| --- | --- |
| Gehler & Nowozin (2009) | 85.5±3.0 |
| BEMKL (one-versus-all) | 85.6±1.2 |
| BEMKL (multiclass) | 85.9±1.2 |

Table 4. Performance comparison on `Flowers102` data set.

| Method | AUC | EER | Accuracy |
| --- | --- | --- | --- |
| Titsias & Lázaro-Gredilla (2011) | 0.952 | 0.107 | 40.0 |
| BEMKL (one-versus-all) | 0.969 | 0.068 | 67.0 |
| BEMKL (multiclass) | 0.969 | 0.069 | 68.9 |

Table 5. Performance comparison on `Caltech101` data set.

| Method | Test Accuracy |
| --- | --- |
| Varma & Babu (2009) | 71.1±0.6 |
| BEMKL (one-versus-all) | 72.7±0.1 |
| BEMKL (multiclass) | 73.1±0.1 |

6.2.1. Protein Fold Recognition Data Set

`Protein` data set considers 27 different fold types of 694 proteins (311 for training and 383 for testing) and it is available at http://mkl.ucsd.edu/dataset/protein-fold-prediction/. There are 12 distinct feature representations summarizing protein characteristics. We construct 12 kernels on these representations and take 20 replications using the experimental procedure of Damoulas & Girolami (2008). Table 2 gives the classification results on `Protein` data set. We see that BEMKL is significantly better than the probabilistic MKL method of Damoulas & Girolami (2008).

6.2.2. Oxford Flowers17 Data Set

`Flowers17` data set contains flower images from 17 different types with 80 images per class and it is available at http://www.robots.ox.ac.uk/~vgg/data/flowers/17/. It also provides three predefined splits with 60 images for training and 20 images for testing from each class. There are seven precomputed distance matrices over different feature representations. These matrices are converted into kernels as $k(\boldsymbol{x}_i, \boldsymbol{x}_j) = \exp(-d(\boldsymbol{x}_i, \boldsymbol{x}_j)/s)$ where $s$ is the mean distance between training point pairs. The classification results on `Flowers17` data set are shown in Table 3. BEMKL achieves higher average test accuracy with smaller standard deviation across splits than the boosting-type MKL algorithm of Gehler & Nowozin (2009). We also see that using the same set of kernel weights for each class as in our multiclass formulation is better than learning separate sets of kernel weights as also observed by Gehler & Nowozin (2009).

6.2.3. Oxford Flowers102 Data Set

`Flowers102` data set contains flower images from 102 different types with more than 40 images per class and it is available at http://www.robots.ox.ac.uk/~vgg/data/flowers/102/. There is a predefined split consisting of 2040 training and 6149 testing images. There are four precomputed distance matrices over different feature representations. These matrices are converted into kernels using the same procedure on `Flowers17` data set. Table 4 shows the classification results on `Flowers102` data set. We report averages of *area under ROC curve* (AUC) and *equal error rate* (EER) values calculated for each class in addition to multiclass accuracy. We see that BEMKL outperforms the GP-based method of Titsias & Lázaro-Gredilla (2011) in all metrics on this challenging task.

6.2.4. Caltech101 Data Set

`Caltech101` data set consists of object images from 102 different categories and it is available at http://www.robots.ox.ac.uk/~vgg/software/MKL/. There are three predefined splits with 15 images for training and 15 images for testing from each class along with 10 precomputed kernel matrices. The classification results on `Caltech101` data set are given in Table 5. BEMKL achieves higher average test accuracy with smaller standard deviation across splits than the discriminative MKL algorithm of Varma & Babu (2009). Similar to the results on `Flowers17` and `Flowers102`, we see that using the same set of kernel weights for each class instead of separate sets is better in terms of generalization performance. This approach allows the classifiers trained for each class to share information with others similar to multitask learning.

## 7. Conclusions

In this paper, we introduce a Bayesian multiple kernel learning framework in order to have computationally feasible algorithms by formulating the combination in a novel way. This formulation allows us to develop fully conjugate probabilistic models and to derive very efficient deterministic variational approximations for inference. We give the inference details for binary classification only due to space limitation and explain briefly how the method can be extended to multiclass learning and semi-supervised learning. Another interesting direction is to formulate a multitask learning method by enforcing similarity between the kernel weights of different tasks.

Experimental results on benchmark binary classification data sets show that the proposed method can combine hundreds of kernels in less than a minute. We also report very good results on one bioinformatics and three image recognition data sets, which contain multiclass classification problems, compared to previously reported results.

**Acknowledgments.** This work was financially supported by the Academy of Finland (Finnish Centre of Excellence in Computational Inference Research COIN, grant no 251170).